# Customers Churn Prediction in Financial Institution Using Artificial Neural Network.

Kamorudeen A. AMUDA*[1], Adesesan B. ADEYEMO[2]

Department of Computer Science, University of Ibadan, Nigeria.

*Email of Corresponding Author: *akindeleamuda@gmail.com*

**ABSTRACT**
In this study a predictive model using Multi-layer Perceptron of Artificial Neural Network architecture was developed to predict customer churn in financial institution. Previous researches have used supervised machine learning classifiers such as Logistic Regression, Decision Tree, Support Vector Machine, K-Nearest Neighbors, and Random Forest. These classifiers require human effort to perform feature engineering which lead to over-specified and incomplete feature selection. Therefore, this research developed a model to eliminate manual feature engineering in data preprocessing stage. Fifty thousand customers' data were extracted from the database of one of the leading financial institution in Nigeria for the study. The multi-layer perceptron model was built with python programming language and used two overfitting techniques (Dropout and L2 regularization).
The implementation done in python was compared with another model in Neuro solution infinity software. The results showed that the Artificial Neural Network software development (Python) had comparable performance with that obtained from the Neuro Solution Infinity software. The accuracy rates are 97.53% and 97.4% while ROC (Receiver Operating Characteristic) curve graphs are 0.89 and 0.85 respectively.

**Keywords:** Multi-layer Perceptron, Artificial Neural Network, Customer Churns, Classifiers, Feature engineering, L2 regularization and Dropout.

## 1.0 Introduction

The volume of data has been growing at a fast rate due to technological advancement which makes data as the oil of the 21$^{st}$ century but oil is just useless until it refines into fuel. Many new methods and techniques have been introduced to extract the meaningful and salient information hidden in the data and the process is called Data Mining Techniques. The technique included Support Vector Machine (SVM), Linear Regression, Genetic algorithm, Decision Tree and Neural Network (Taiwo & Adeyemo 2012).

Customer churn is a fundamental problem for companies and it is defined as the loss of customers because they move out to competitors. Being able to predict customer churning behavior in advance, gives an institution a high valuable insight in order to retain and increase their customer base.

A wide range of customer churn predictive models have been developed in the last decades. Most advanced models make use of state-of-the-art machine learning classifiers such as random forest, linear and logistic regression (Castanedo, 2014).

One of the most direct and effective approaches to keep the current customers is that the company should be able to foresee potential churners in time and react to it quickly. Recognizing the indications of potential churn; satisfying customer needs, restoring and re-establishing loyalty are actions supposed to help the s organization minimize the cost of gaining new customers (Mitkees et al, 2017).

Financial institutions are the organization that process monetary transaction including business and private loans, customer deposits and investments. As customers are directly related to profits, financial institutions must avoid the loss of customers while acquiring new customers. Harvard Business Review believes that by reducing the customer defection rate by 5%, companies can increase profits by 25% to 85%, while Business Week thought the profits will increase by 140%.

Bhattacharya (1994) that the cost of developing a new customer is 5 to 6 times than retaining an old customer. As can be seen, reducing customer attrition has a significant impact not only on increasing profits for financial institutions, but also on enhancing their core competitiveness. Therefore, it is urgent for financial institutions to improve the capabilities to predict customer churn, thereby taking timely measures to retain customers and preventing other clients from churning.

Artificial Neural Network also called Neural Network (NN) is a complex network that comprises a large set of simple nodes known as neural cells. Artificial Neural Network was proposed based on advanced biology research concerning human brain tissue and neural system, and can be used to simulate neural activities of information processing in the human brain (Vaxevanidis,2008).

Artificial Neural Network have been used in finance for portfolio management, credit rating and predicting bankruptcy, forecasting exchange rates, predicting stock values, inflation and cash forecasting.

In this paper, we work on predictive analysis of churning behavior in financial institution base on the dataset extracted from the bank database. The model developed with MLP of Artificial Neural Network architecture, potential churners and non-churners were predicts and different performance metrics such as precision, accuracy, recall and f-score were used to evaluate the robustness of the model and compared with Neuro Solution Infinity analytical tool. The paper is structured as follows. Section 2 explores the various literature reviews on the related work. Section 3 gives the research objectives and illustrates the methodological approach. Section 4 illustrates the results of the proposed work and Section 5 explores the conclusion and future work.

## 2.0 LITERATURE REVIEWS

He et al, (2014) carried out a research on customer attrition analysis of commercial bank using Support Vector Machine. The dataset contains 50000 customers records were extracted from Chinese commercial bank data warehouse. After the removal of missing values and outliers, 46,406 records used to model. SVM algorithm was applied but due to the imbalanced characteristics of the dataset, Random sampling method was introduced to improve SVM since it has a higher degree of recognition. The results shown that the combination of random sampling and support vector machine algorithm significantly improved the predictive power and accurately predict churning rate.

In the telecommunication industry, cellular network providers are becoming more competitive and churn management has become a crucial task telecommunication industry. Sharma, (2011) applied a neural network to predict customer churn in cellular network service. The dataset contains 20 variables worth of information about 2,427 customers were collected from UCI Machine Learning Database at the University of California, Irvine. Neural Network was implemented on Clementine data mining software package from SPSS Inc., Clementine provided two different classes of supervised neural networks namely Multilayer Perceptron (MLP) and Radial Base Function Network (RBFN). The result shows that the model predicts customers churn with an accuracy rate of more than 92%.

Oyeniyi & Adeyemo, (2015) customer churn has become a major problem within a customer-centered banking industry and banks have tried to track Customer Interaction in order to detect early warning signs in customer's behaviour such as a reduction in transaction and account dormancy. They also worked on customer churn analysis in the banking sector, the model used K-Means and Repeated Incremental Pruning to Produce Error Reduction (JRip algorithm) which was implemented on Weka. The dataset was extracted from the bank's customer relationship management database and transaction warehouse from a major Nigeria bank. The results determine the pattern in customer behavior and help banks to identify customers that are likely to churn.

To improve the prediction abilities of machine learning method, Guo-en et al (2008) applied support vector machine on structure risk minimization to customer churn analysis. The dataset was collected from machine learning UCI database of the University of California. The results shown that the support vector machine method enjoys the best accuracy rate, hit rate, covering rate and lift coefficient when compared with Artificial Neural Network, Decision Tree, Logistic Regression and Naïve Bayesian classifier. Support vector machine provides an effective measurement for customer churn prediction.

Taiwo & Adeyemo (2012), used descriptive and predictive data mining techniques to determine the calling behaviour of subscribers and recognize subscribers with a high probability of churn in a telecommunication company. In the descriptive stage, the customers were clustered based on their usage behavioural features and algorithms used for clustering methods were K-Means and Expected Maximization (EM). In the predictive stage, DecisionStump, M5P and RepTree classifiers algorithms were implemented in Weka. The results show that EM performs better than K-mean in the descriptive stage while M5P perform better than both DecisionStump, and RepTree in the predictive stage.

Lu et al, (2011) proposed the use of boosting algorithm to predict the customer churn in the telecommunication industry. Customers were separated into two clusters based on the weight assigned by boosting algorithm, the data set used was extracted from a telecommunication company which included a segment of mobile customers who are active and contains about 700 variables. Boosting algorithm performs better than logistic regression and it provides a good separation of churn data.

Sandeepkumar & Monica (2019), considered multi-layered neural network which also known as Deep Feed Forward Neural Network (DFNN) to perform predictive analytics on customer attrition in the banking sector. The dataset was collected from the

UCI machine learning archive which has total 10,000 customer data with 14 dimensions of features. The model use optimized one hot encoding and Tukey outliers' algorithms to perform data cleaning and preprocessing. The model was compared with the machine learning algorithms such as Logistic regression, Decision tree and Gaussian Naïve Bayes algorithm. The results shown that Enhanced Deep Feed Forward Neural Network (DFNN) model performs best in accuracy when compared with others machine learning algorithms.

Wang et al (2018), worked on large-scale ensemble model for customer churn prediction in search Ads. The aimed of the research was to detect customers with a high propensity to leave the ads platform. the ensemble model of gradient boosting decision tree (GBDT) was used to predict customer that will be a churner in the foreseeable future based on its activities in the search ads. two different features for the GBDT were dynamic features and static features. dynamic features considered a sequence of customers' activities such as impressions, clicks during a long period. while static features considered the information of customers setting such as creation time, customer type. the dataset was collected from Bing Ads platform and the result shown that the static and dynamic features are complementary with AUC (area under the curve of ROC) value 0.8410.

Rosa (2019), proposed a new framework that used Artificial Neural Network for assessing and predicting customer attrition in the banking industry. About 1588 customers data from the period of January 2017 to December 2017 was extracted from the bank´s Data Warehouse with the use of SAS Base. The research solely focused on developing neural network and overlooked other machine learning algorithms such as Decision Trees, Logistic Regression, or Support Vector Machines.

## 3.0 MATERIALS AND METHODS

Customer churn prediction has a great impact on financial institutions since they are mainly depending on customer satisfaction for their operations. These institutions are highly competitive environment and retain customers by satisfying their needs under the resource's constraint. The data mining techniques used to discover interesting patterns and relationships that exist in the data and predict the customers' behavior either churning or non-churning by fitting the model on the available historical data.

Researches on churn prediction have used machine learning classifiers such as Logistic Regression, Decision Tree, Support Vector Machine, K-Nearest Neighbors, and Random Forest. These classifiers require human effort to perform feature engineering which lead to over-specified and incomplete feature selection.

Hence, this research developed a predictive model using Multilayer Perceptron of Artificial Neural Network architecture to predict customer churn as well as to eliminate the manual feature engineering process in data preprocessing stage.

The main objectives of this research paper are: To develop customer churn predictive model using Multilayer Perceptron of Artificial Neural Network Architecture; To use the model to predict the potential churners and non-churners; To evaluate and compare the performance of the model with analytical tool result.

**Data Collection**

The dataset was extracted from the database of one of the leading financial institution in Nigeria. The contained about 50000 customers' data with 42 attributes. Some of these are:

| Features | Description |
| --- | --- |
| Gender | M for Male; F for Female and NULL was used for customers we couldn't get their gender type |
| Cust_txn_status | Active and Inactive |
| Marital_status | D= Divorced; M= Married; S=Single, NULL=NULL |
| Occupation | Customer Job description |
| Lga | Local Government Area |
| State | State in Nigeria |
| Religion | Christian; Islam; Other Religion |
| current_account | Current Accounts for individuals |
| xclusive_subscript | Current Account for High Net worth Individuals with monthly income above N1m |
| current_account_corp | Current account for big companies like shell, chevron, coca-cola, nestle. |
| savings_deposit_youth | Savings accounts for teenagers and babies |
| Community_savings_account | Savings account for estates and communities |
| Hida | High interest deposit account |
| Mobapp_Fund_Trsf_LCY_Revenue | Profit accruable to the Bank based on currency transfers by the customer between Sept to November |
| Mobapp_Fund_Trsf_FCY_Count | The total number of times the customer transferred money (USD or EURO or GBP only). |
| Mobapp_Fund_Trsf_FCY_Vol | The total amount of money the customer transferred (USD or EURO or GBP only) |
| Mobapp_Fund_Trsf_FCY_Revenue | Profit accruable to the Bank based on foreign currency transfers made by the customer between Sept to November |
| Mobapp_Lifestyle_Count | The total number of times the customer paid for travel tickets or movie tickets or other event ticket |
| Mobapp_Lifestyle_Vol | Amount in naira times paid by customer for travel tickets or movie tickets or other event ticket. |
| mobapp_Lifestyle_Revenue | Profit accruable to the bank based on transactions made by customers for travel tickets or movie tickets or other event. |
| Cust_id | Unique identification number for customers. |

## Data Preparation

The missing values with 30% null were removed from the dataset with the aid of Python programming language libraries. Numerical data was replaced with the 'mean' of the variables while the 'mode' was used for the categorical data. To achieve better performance, the categorical data was transformed to numerical format using the Label Encoder function in Python. Feature scaling was applied to normalize the data and improved the computational time.

## Tools and Libraries

**Anaconda:** is an open source software distribution of R and Python programming languages that are used for scientific computing such as data science, predictive analytics, machine learning, and deep learning applications purposely to simplify package management and deployment.

**Jupyter Notebook**: is an open free source web application that is used for data cleaning and transformation, numerical, simulation, statistical modelling, data visualization and so on.

**Matplotlib:** is an amazing visualization library in Python programming language for two-dimensional plots of arrays. One of the greatest benefits of visualization is that it allows high dimensionality data to visualize and easily understandable and it consists of several plots like line, bar, scatter, histogram and so on.

**Pandas:** is the most popular Python programming language package that offers powerful, expensive and flexible data structures that make data manipulation and analysis easy.

**Numpy:** is the fundamental package for scientific computing in Python programming language that contains a powerful N-dimensional array object and also useful in linear algebra.

**Label Encoder:** is a Python programming language package that is used to transform non-numerical labels (or nominal categorical variables) to numerical labels.

**ANN visualizer**: is a Python programming language library that enables visualization of an artificial neural network.

**Seaborn:** is a Python data visualization library based on matplotlib that provides high-level interface for drawing attractive and informative statistical graphics.

**Sci-kit Learn**: it is a free machine learning library for Python programming language that designed to interoperate with Python numerical Numpy and scientific libraries SciPy. Also, it can be used for classification, regression and clustering algorithms including support vector machine, linear and logistic regression, random forests, gradient boosting, decision tree, K-means and so on.

**TensorFlow**: is an open source Python library used for machine learning applications such as neural network and used Keras as a backend.

**Keras**: is a neural network framework for Python programming language that provides a convenient way to define and train almost any kind of deep learning model.

**Neuro Solutions Infinity:** is one of the most powerful neural network software of the Neuro Solutions family that streamlines the data mining process by automatically cleaning and preprocessing data. It uses distributed computing, advanced neural networks and artificial intelligence (AI) to model data; creates highly accurate predictive models with an easy-to-use and intuitive interface that provides valuable insights that can be used to drive better decisions.

**Microsoft SQL Server:** is a relational database management system developed by Microsoft Incorporation which primarily used for storing and retrieving data requested by other software applications that run on the same computer or on another computer across a network.

**Flowchart**: illustrate the step by step procedure on how to predict customers churn with the multilayer perceptron of artificial neural network architecture.

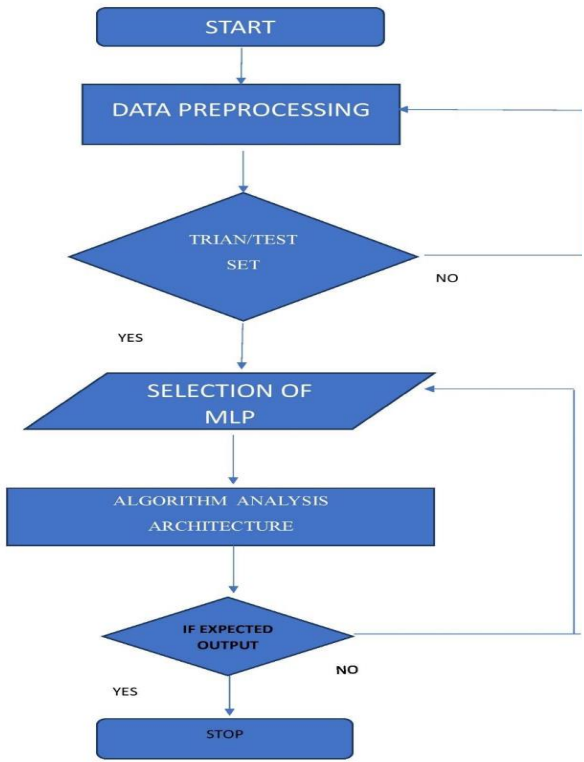

**Figure 1: Flow Diagram for proposed model.**

### Implementation

The implementation was carried out in two phases; MLP software development on Python and then using Neuro Solution Infinity software. Multilayer Perceptron: MLP is an architecture of the artificial neural network that consists of multiple layers where each layer is fully connected with the next layer in a feed-forward direction. The first layer and the last layers represent the inputs (independent variables) and outputs (target variables) of the system respectively. Connections between the nodes represented as weights. The more the numbers of hidden layers in the network, the more the complexity of the network.

In multilayer perceptron architecture, each hidden layer node consists of two parts;

- Summation function: The summation function calculates the sum of each input value multiplied by the corresponding weights. Mathematically, it can be represented as

$$S = \sum_{i=0}^{n} W_i X_i \qquad (1)$$

Where n is the numbers of neuron in the network

- Activation function: The activation is applied by each neuron to sum of weight input signals in order to determine its output signal. Sigmoid function such as the S-shaped curve is one of the most commonly used activation functions for binary classification problems.

$$\emptyset(S) = \frac{1}{1 + \exp(-S)} \qquad (2)$$

**Steps for MLP Model**

i. Extraction of customer data including transactional history from bank database.
ii. Data pre-processing for removal of noise and transformation of categorical data into numerical data.
iii. Data splitting (train and test set) and feature scaling
iv. Start at the input layer by forwarding propagates patterns of the train data through the network to generate output.
v. Using cost function to calculate the network output to minimize the error rate.
vi. Find the derivative with respect to each weight in the network and updating the model,
vii. Calculate the network output and apply threshold function to obtain the predicted class label.
viii. Evaluate the model.

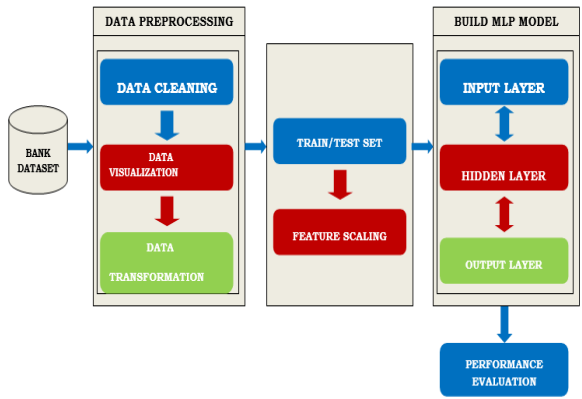

**Figure 2: Proposed Model of Multilayer Perceptron of Artificial Neural Network.**

Neuro Solution Infinity Software: The implementation was done by launching the software followed by uploading the data from the Personal Computer (PC) into the software platform. The dataset divided into three parts; training set, test set, and validation set and the threshold for the missing values was set. The Neuro Solution Software comprises of different algorithms such as Multilayer Perceptron (MLP), Probabilistic Neural Network (PNN), and Support Vector Machine (SVM). In figure 3 about 50 different models ran on the software and the best model was selected for the churn analysis.

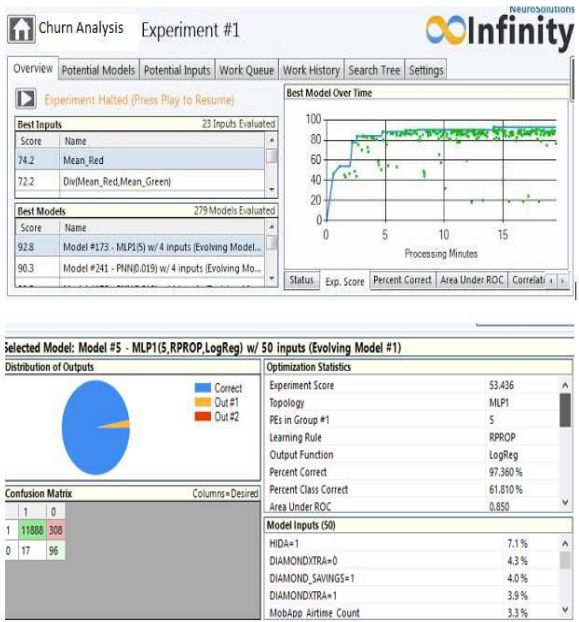

**Figure 3: Screenshot of Neuro Solution Infinity Interface**.

## 4.0 RESULTS AND DISCUSION

The result of the implementation in Python was presented in figure 4, figure 5 and figure 6. It can be seen that training loss was decreasing while training accuracy was increasing and also the validation loss was decreasing while validation accuracy was increasing. These shown that the model was neither overfitting nor underfitting.

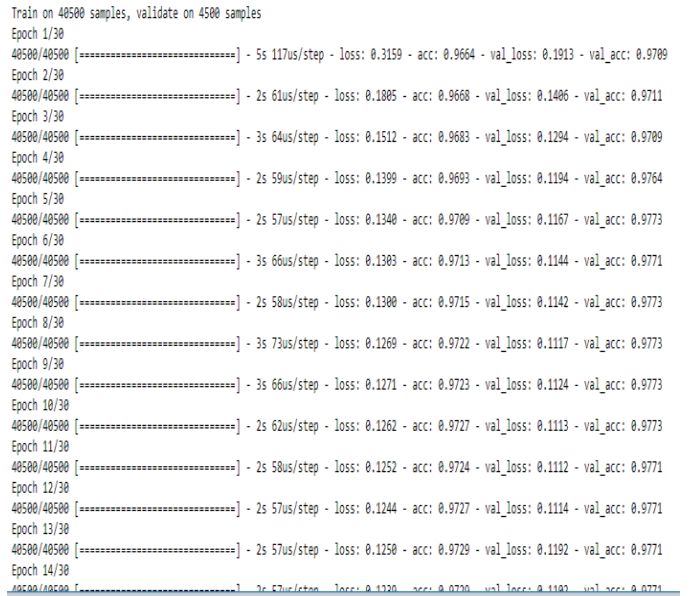

**Figure 4: Learning Process**

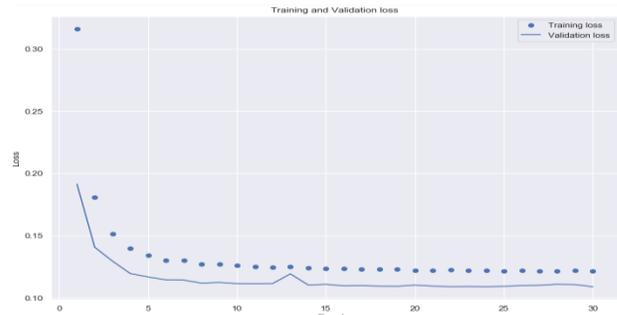

**Figure 5: Training Loss and Validation Loss**

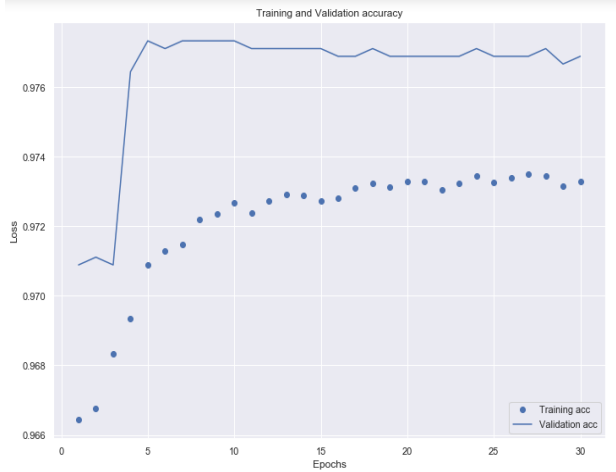

**Figure 6: Training Accuracy and Validation Accuracy**

**Performance Evaluation:** Confusion matrix was used to describe the performance of the Multilayer perceptron classifier on a set of test data for which the true values are known. It aids the visualization of the performance of the model. The confusion matrix table summarized the prediction results of the model. The numbers of correctly classified and incorrectly classified (misclassified) are summarized with count values and broken down by each class (churners and non-churners). The essence of performance evaluation is to gives insight not only into the errors being made by the model but more importantly the types of errors that are being made.

ROC curve was used to know the capability of the models to distinguish between classes (churners and non-churners). ROC values range between 0 and 1, that is, if the value close to 1, it means the model has good measure of separability while near 0 means it has worst measure of separability (i.e. it is reciprocating the result). Figure 7 it can be seen that the model correctly classified 38 customers as churners, 4841 as non-churners and incorrectly misclassified 9 customers as churners and 112 as non-churners.

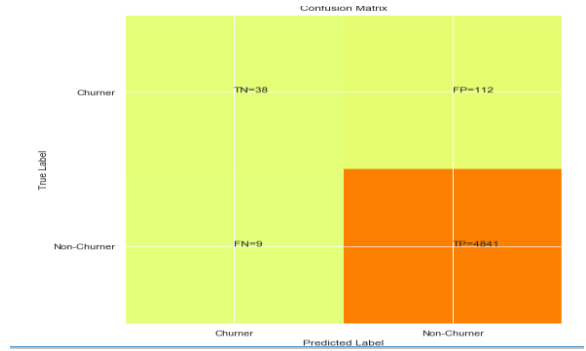

**Figure 7 Confusion Matrix on Python**

From figure 8 the value for ROC curve graph was 0.89 which mean the model was working perfectly. Dropout was introduced in the network to reduce overfitting and the complexity of the network to the minimal level.

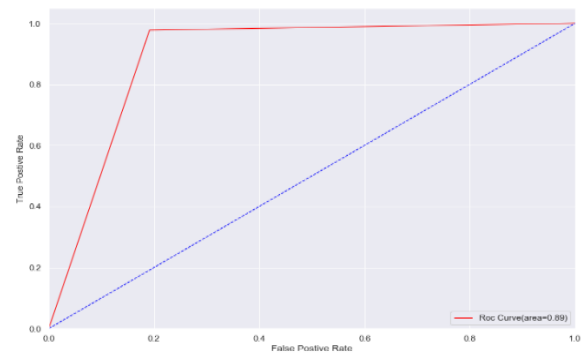

**Figure 8: ROC Curve on Python**

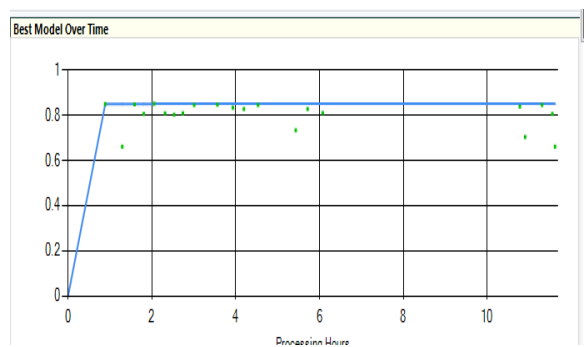

**Figure 9: Neuro Solution Infinity ROC Curve**

Figure 9 shows the ROC curve for the best model in the Neuro Solution Infinity software, Multilayer Perceptron (MLP) model gave the best result with 0.85 followed by Probabilistic Neural Network (PNN), 0.84 and Support Vector Machine (SVM), 0.842.

## Comparative Analysis of Implementations

Table 1 presents the results of the two implementations; it can be seen that the Multilayer Perceptron implemented on Python's result was comparable to that of the Neuro Solution Infinity software.

|  | Python | Neuro Solution Infinity Software |
|---|---|---|
| Accuracy Rate | 97.53 | 97.36 |
| ROC curve Value | 0.89 | 0.85 |
| Precision | 97.7 | 97.6 |
| Recall | 99.8 | 99.9 |
| F – Measure | 98.8 | 98.7 |

**Table 1: Implementation Results**

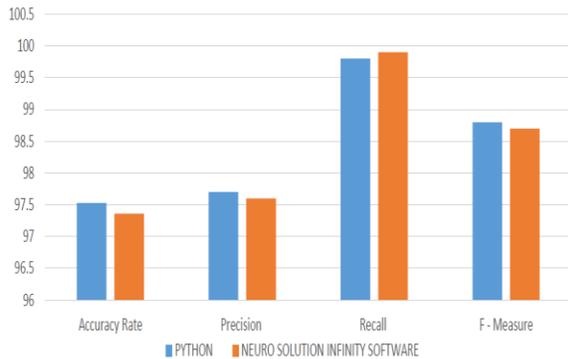

**Figure 10:** Pictorial view of the two implementations performance.

In figure 11 Dropout was introduced in the network to reduce the degree overfitting and the complexity of the network to the minimal level.

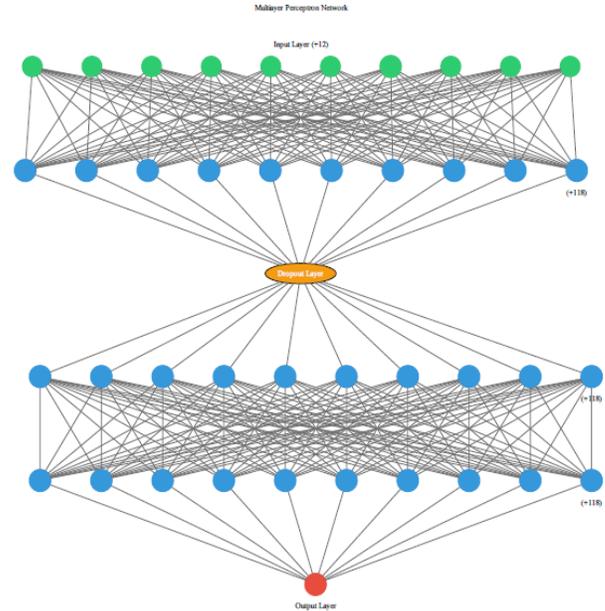

**Figure 11:** Schematic Diagram of the Proposed Multilayer Perceptron (MLP) model

## 5.0 CONCLUSION

This research was designed for the prediction of customers churn using the data from a financial institution in Nigeria. The data was extracted from the bank database and divided into three sets: training set, test set, and validation set. Eighty percent (80%) of the dataset was used for training, ten percent (10%) was used for testing and the remaining ten percent (10%) was used for validating the model. Data pre-processing was carried out by removing thirty percent (30%) of missing values and the text (categorical) data were converted into numerical data. Also, feature scaling was employed to increase the computation time of the algorithm.

Multilayer Perceptron Artificial Neural Network architecture was implemented on two different platforms; Python and Neuro Solution Infinity Software. The results obtained using the two software were comparable with the Python implementation given a 97.53 % accuracy while the Neuro Solution Infinity software gave 97.36% accuracy. The performance of the models was evaluated on the test dataset (unseen data) using the following metrics: Accuracy, Precision, Recall, F-measure. In recent surveys Python programming language has been gaining in popularity as the tool of choice amongst Data Scientists and Data Mining and Natural Language researchers. This is due to its being relatively easy to learn and its very rich library of functions and utilities. The information obtained from the predictive model can be used for decision making in customer retention management system. For future work other ANN architectures such as Convolution Neural Network (CNN), Recurrent Neural Network (RNN) and Long Short-Term Memory (LSTM).

**REFRENCES**

Bhattacharya, C. B. (1994). When Customers Are Members: Customer Retention in Paid Membership Contexts?

Castanedo, F. (2014). Using Deep Learning to Predict Customer Churn in a Mobile Telecommunication Network, 1–8.

Guo-en, X. I. A., & Wei-dong, J. I. N. (2008). Model of Customer Churn Prediction on Support Vector Machine. *Systems Engineering - Theory & Practice*, *28*(1),71–77. https://doi.org/10.1016/S1874-8651(09)60003-X

He, B., Shi, Y., Wan, Q., & Zhao, X. (2014). Prediction of customer attrition of commercial banks based on SVM model. *Procedia - Procedia Computer Science*,*31*,423–430. https://doi.org/10.1016/j.procs.2014.05.286

Lu, N., Lin, H., Lu, J., & Zhang, G. (2011). A Customer Churn Prediction Model in Telecom Industry Using Boosting, (c), 1–7.

Mitkees, I. M. M., Ibrahim, A., & Elseddawy, B. (2017). Customer Churn Prediction Model using Data Mining techniques, 262–268.

Oyeniyi, A. O., & Adeyemo, A. B. (2015). Customer Churn Analysis in Banking Sector Using Data Mining Techniques, *8*(3), 165–174.

Rosa, Nelson Belém da Costa. *Gauging and foreseeing customer churn in the banking industry: a neural network approach*. Diss. 2019

Sandeepkumar hegde, & Mundada, M. R.(2019). *Enhanced Deep Feed Forward Neural Network Model for the Customer Attrition Analysis in Banking Sector*. (July), 10–19. https://doi.org/10.5815/ijisa.2019.07.02

Sharma, A. (2011). A Neural Network based Approach for Predicting Customer Churn in Cellular Network Services, *27*(11), 26–31.

Taiwo, Adeyemo. (2012). Computing, Information Systems Data Mining Technique for Predicting Telecommunications Industry Customer Churn Using both Descriptive and Predictive Algorithms Customer Churn Using both Descriptive and Predictive Algorithms, *3*(2), 27–34.

Wang, Q., Xu, M., & Hussain, A. (2018). *Large-scale Ensemble Model for Customer Churn Prediction in Search Ads*.

Vaxevanidis, N. M. (2008). Artificial neural network models for the prediction of surface roughness in electrical discharge machining, 283–292. https://doi.org/10.1007/s10845-008-0081-9